\documentclass[journal]{IEEEtran}

\usepackage{cite}
\usepackage{amsmath,amssymb,bm}
\usepackage{array,booktabs,multirow,tabularx}
\usepackage{graphicx}
\usepackage[table,dvipsnames]{xcolor}
\usepackage{tikz}
\usetikzlibrary{arrows.meta,fit,positioning}
\usepackage[hidelinks]{hyperref}
\usepackage{url}

\newcolumntype{L}[1]{>{\raggedright\arraybackslash}p{#1}}
\newcolumntype{C}[1]{>{\centering\arraybackslash}p{#1}}
\newcolumntype{Y}{>{\raggedright\arraybackslash}X}
\newcolumntype{Z}{>{\centering\arraybackslash}X}

\newcommand{\Supported}{\cellcolor{ForestGreen!10}\textcolor{ForestGreen!65!black}{\bfseries\checkmark}}
\newcommand{\Unsupported}{\textcolor{gray!70}{--}}
\newcommand{\PHYDim}{\cellcolor{NavyBlue!8}\textcolor{NavyBlue}{\bfseries PHY}}
\newcommand{\RANDim}{\cellcolor{TealBlue!8}\textcolor{TealBlue!80!black}{\bfseries RAN}}
\newcommand{\ISACDim}{\cellcolor{BurntOrange!10}\textcolor{BurntOrange!85!black}{\bfseries ISAC}}

\DeclareMathOperator{\NMSE}{NMSE}
\DeclareMathOperator{\SGCS}{SGCS}
\begin{document}

\title{CFM-Bench: A Unified Multi-Domain, Multi-Task Benchmark for Channel Foundation Models}

\author{Yuan Gao, \IEEEmembership{Member,~IEEE},
Wenjun Yu,
Jun Jiang,
Yunfan Li,
Xinyu Guo,
and Shugong Xu, \IEEEmembership{Fellow,~IEEE}%
\thanks{This work was supported in part by the Shanghai Natural Science Foundation under Grant 25ZR1402148 and in part by the 6G Science and Technology Innovation and Future Industry Cultivation Special Project of the Shanghai Municipal Science and Technology Commission under Grant 24DP1501001. (Shugong Xu is the corresponding author.)}%
\thanks{Yuan Gao, Wenjun Yu, Yunfan Li, and Xinyu Guo are with the School of Communication and Information Engineering, Shanghai University, Shanghai, China (e-mail: gaoyuansie@shu.edu.cn; yuwenjun@shu.edu.cn; lyf2023@shu.edu.cn; guoxinyu@shu.edu.cn).}%
\thanks{Jun Jiang and Shugong Xu are with Xi'an Jiaotong-Liverpool University, Suzhou, China (e-mail: jun.jiang25@student.xjtlu.edu.cn; shugong.xu@xjtlu.edu.cn).}}

\maketitle

\begin{abstract}
Channel foundation models (CFMs) are commonly evaluated in model-specific pipelines that differ in data, radio configurations, partitions, adaptation procedures, task definitions, and metrics, preventing reproducible comparison across CFMs and against task-specific networks. We release CFM-Bench, a unified multi-domain, multi-task benchmark comprising 157,900 official single-frame examples from six domains spanning 3GPP statistical simulation, two ray-tracing pipelines, terrestrial and aerial measurements, and synchronized vehicular multimodal simulation. Source-specific interfaces preserve complex channel state information (CSI) and the physical metadata available in each domain while allowing documented model-specific preprocessing. To reduce spatio-temporal leakage, official partitions isolate complete trajectories, measurement sessions, flights, vehicle links, simulation realizations, or buffered spatial regions. CFM-Bench excludes all benchmark splits from foundation-model pretraining, reserves the official training split for downstream fine-tuning, and reports the data used during model development. Six task groups across physical-layer, radio-access-network, and integrated sensing and communication applications cover CSI feedback, frequency and temporal channel extrapolation, propagation-state classification, current- and future-beam prediction, and single-frame and temporal localization. Representative experiments on CSI feedback, channel extrapolation, current-beam prediction, and wireless positioning provide reproducible reference results for pretrained channel-prediction models and task-specific neural networks. These results show that relative model performance can vary across data domains, highlighting the importance of using common data partitions, task definitions, and evaluation metrics. CFM-Bench provides a common substrate for evaluating the transferability of channel representations across models, domains, and tasks.
\end{abstract}


\IEEEpeerreviewmaketitle
\bstctlcite{IEEEexample:BSTcontrol}

\section{Introduction}
\label{sec:introduction}
Artificial intelligence (AI) is moving from a support tool for network tuning toward a native component of the wireless air interface. The 3rd Generation Partnership Project (3GPP) has studied how AI/machine-learning (AI/ML) can be integrated into the New Radio (NR) air interface, including data collection, model training, adaptation, inference, and lifecycle management~\cite{3gpp38843,gao2026ai}. In parallel, foundation models have shown that one pretrained representation can serve many downstream tasks after light adaptation~\cite{bommasani2021opportunities}. Applied to wireless channels, this idea motivates channel foundation models (CFMs): models that learn reusable representations from channel state information (CSI), impulse responses, geometry, and related context~\cite{jiang2025towardscfm}.

CFM research is advancing quickly. Recent work uses masked reconstruction, autoencoders, and related self-supervised objectives for channel reconstruction, prediction, and other downstream tasks~\cite{alikhani2025lwm,liu2025wifo,jiang2026csimae}. Most reported gains, however, come from isolated experimental pipelines. Each study uses its own data and protocol and typically compares a pretrained model with a supervised baseline only within that setting. Because pretraining corpora, radio configurations, data partitions, task definitions, adaptation budgets, and metrics differ across papers, published scores neither establish a stable ranking among CFMs nor show whether a general-purpose representation outperforms a strong task-specific network under matched conditions.

Current CFM evaluation practice has three recurring limitations.
\begin{itemize}
    \item First, even when several CFMs are compared in one study, their broader results remain tied to different pretraining data, system configurations, partition schemes, task definitions, and metrics. The community therefore lacks a common downstream protocol for assessing transferability or comparing CFMs fairly with task-specific architectures.
    \item Second, most studies train and evaluate within a single channel-generation mechanism. Statistical models stress distributional variation, ray-tracing environments impose geometry-consistent multipath, measured channels expose hardware and calibration effects, and multimodal settings add sensing context. Evaluation on only one mechanism leaves robustness across these physical regimes largely untested.
    \item Third, wireless channels are strongly correlated in space and time. Samples from the same trajectory, measurement session, vehicle link, or local spatial region often share similar multipath structure, angles, and delays. Random frame-level splits can therefore place near-duplicate channels on both sides of the split, inflate reported performance, and conceal overfitting.
\end{itemize}

A meaningful benchmark must consequently provide more than another study-specific score table. It should separate external pretraining data from official downstream partitions, reduce spatio-temporal leakage, document the data used during model development, and expose models to complementary propagation mechanisms. At the same time, it should preserve domain-specific physical semantics rather than forcing heterogeneous channels into superficially uniform tasks.

CFM-Bench is built for this purpose. Rather than generating another single-source dataset, it curates one representative configuration from each of six existing sources and organizes them within a common evaluation framework. The resulting domains cover 3GPP statistical simulation, two independent ray-tracing pipelines, industrial and aerial measurements, and synchronized vehicular multimodal simulation. Source-specific interfaces retain complex CSI and available physical metadata, while official partitions, task definitions, metrics, and reporting conventions provide a shared downstream comparison substrate for CFMs and task-specific networks.

The main contributions are as follows.
\begin{itemize}
    \item \textbf{Matched evaluation objective.} We formulate CFM evaluation as comparison under common downstream partitions, observations, targets, metrics, and adaptation rules, enabling comparisons among CFMs and against task-specific baselines.
    \item \textbf{Multi-mechanism test conditions.} We curate six domains spanning statistical variation, geometry-consistent multipath from independent ray-tracing pipelines, terrestrial and aerial measurements, and synchronized multimodal context, while preserving their native channel structures and metadata.
    \item \textbf{Leakage-resistant partitioning.} We construct partitions from complete simulation realizations, spatial regions, trajectories, measurement sessions, flights, or vehicle links, and define a data-use protocol that separates foundation-model pretraining from downstream adaptation and test evaluation.
    \item \textbf{Domain-aware tasks and baselines.} We define six PHY, RAN, and ISAC task groups with validity masks, horizons, codebooks, and coordinate frames appropriate to each domain. Numerical and semantic validation determines the supported domain--task pairs, while representative experiments on CSI feedback, channel extrapolation, current-beam prediction, and positioning establish reproducible reference points.
\end{itemize}

The remainder of the paper is organized as follows. Section~\ref{sec:related} reviews channel datasets, CFMs, and the unresolved evaluation gap. Section~\ref{sec:design} translates that gap into benchmark design choices, domain curation, leakage-resistant partitions, and quality-control procedures. Section~\ref{sec:protocols} defines the comparison methodology and the PHY, RAN, and ISAC tasks, after which Section~\ref{sec:experiments} reports representative baselines. Sections~\ref{sec:availability} and~\ref{sec:limitations} discuss data availability and limitations, followed by the conclusion in Section~\ref{sec:conclusion}.

\section{Related Work}
\label{sec:related}

\subsection{Channel Datasets and Generation Platforms}
\label{sec:related-data}

DeepMIMO introduced a configurable interface for generating massive-MIMO channels from ray-tracing scenarios, enabling reproducible learning studies on spatially consistent channels~\cite{alkhateeb2019deepmimo}. Sionna and Sionna RT later offered open, differentiable tools for physical-layer simulation and radio propagation~\cite{hoydis2022sionna,hoydis2023sionnart}. These platforms allow large-scale channel generation, yet dependence on a single simulator can tie model performance to simulator-specific geometry, material assumptions, and data formatting conventions.

Several public datasets complement these platforms. MOCSID supplies Sionna-RT-generated pedestrian trajectories under overlapping coverage from ten outdoor base stations~\cite{makhlouf2025mocsid,makhlouf2025mocsiddata}. DICHASUS provides phase-coherent measured massive-MIMO channels together with accurate position labels; its ADXX release records a distributed 64-antenna deployment in an industrial setting~\cite{euchner2021dichasus,euchner2024dichasusadxx}. The MaMIMO-UAV campaign captures wideband air-to-ground channels between an $8\times8$ base-station array and a moving UAV~\cite{cui2023mamimouav,colpaert2023mamimouavdata}. Multimodal-Wireless aligns ray-traced CSI with camera, depth, LiDAR, radar, inertial, and positioning data inside CARLA environments~\cite{mao2026multimodalwireless}. Each resource serves its original purpose well. CFM-Bench retains the provenance of these sources while adding shared partitions and task protocols that enable cross-domain evaluation.

\subsection{Channel Foundation Models}
\label{sec:related-cfm}

Current CFM studies can be grouped into two broad methodological lines~\cite{jiang2025towardscfm}. One line focuses on learning reusable representations from unlabeled channel observations. LWM employs masked channel modeling to pretrain a task-agnostic feature extractor~\cite{alikhani2025lwm}. WiFo learns transferable space--time--frequency structure specifically for channel prediction~\cite{liu2025wifo}. CSI-MAE extends masked autoencoding to support both channel perception and generation tasks~\cite{jiang2026csimae}. CSI-CLIP++ aligns frequency-domain CSI with its delay-domain impulse response through contrastive pretraining, targeting channel identification, beam prediction, and positioning~\cite{jiang2026csiclippp}. Filter-and-Attend instead emphasizes robustness when observed CSI is corrupted by noise and interference~\cite{wang2026filterattend}.

A second line expands the task or system-configuration scope of a single pretrained model. The wireless multi-task model in~\cite{sheng2025wirelessfm} jointly handles CSI, location, and traffic prediction within a unified sequence formulation. WiFo-E targets end-to-end frequency-division-duplex precoding and employs sparse mixture-of-experts routing to accommodate varying antenna and user configurations~\cite{wen2026wifoe}. These approaches reduce the need for a separate network at every operating point, yet their downstream protocols remain closely tied to the individual model and application.

Together, these studies illustrate the promise of reusable channel representations and task- and configuration-aware adaptation. Their reported results, however, remain difficult to compare directly. Pretraining corpora, channel representations, available modalities, data partitions, adaptation budgets, target definitions, and evaluation metrics differ across works. CFM-Bench complements ongoing model development by supplying a fixed evaluation substrate with test-unit isolation and explicit reporting of pretraining and adaptation data. In this way it enables transfer assessment under matched downstream conditions without prescribing any particular CFM architecture or pretraining objective.

\subsection{Evaluation Practice and Open Gaps}
\label{sec:related-gaps}

Existing datasets and models address complementary parts of the CFM evaluation problem, but their protocols remain model-centric. Dataset releases generally preserve the partitions and labels needed for their original applications, whereas CFM papers adapt those resources through model-specific preprocessing, observation budgets, and scoring conventions. As a result, a performance difference can reflect the propagation source, the split, the available metadata, the adaptation budget, or the metric as much as the learned representation itself. Random frame-level evaluation further obscures the distinction between interpolation among correlated samples and generalization to an independent physical unit.

The remaining gap is therefore not simply a shortage of channel data or model architectures. What is missing is a shared downstream methodology that combines multi-mechanism coverage, leakage-resistant partitions, physically valid task definitions, and transparent reporting of data exposure. This observation motivates the benchmark design developed next: heterogeneous sources are retained for their complementary physics, while a common evaluation and reporting methodology treats their differences as explicit experimental dimensions rather than hidden confounders.

\section{Benchmark Design}
\label{sec:design}

This section translates the evaluation gaps identified above into concrete benchmark design choices. It first states the evaluation goal and supporting design principles, then introduces the multi-mechanism data domains, the unit-level partitioning strategy and validity checks, and the data interfaces used by downstream models. Task definitions and the detailed evaluation methodology follow in Section~\ref{sec:protocols}.

\subsection{Evaluation Goal and Design Principles}
\label{sec:design-goals}

CFM-Bench is organized around one evaluation goal and three supporting design principles.
\begin{enumerate}
    \item \textbf{Evaluation goal.} Downstream evaluations use common partitions, observations, target definitions, and metrics, enabling CFMs to be compared with one another and with task-specific models under matched experimental protocols.
    \item \textbf{Test conditions.} The evaluation domains span complementary propagation mechanisms and physical contexts rather than many closely related configurations from a single simulator.
    \item \textbf{Partitioning rules.} Split construction accounts for spatio-temporal channel correlation and separates external pretraining data from the official downstream partitions.
    \item \textbf{Task validity.} Task definitions follow the channel semantics, temporal continuity, codebooks, coordinate frames, and available metadata of each source; domain--task pairs without physically meaningful supervision are omitted.
\end{enumerate}
The latter three principles make the evaluation goal operational: mechanism diversity provides complementary test conditions, partitioning rules limit information leakage, and domain-aware validation preserves the physical meaning of the reported scores.

Each source therefore contributes one fixed combination of environment, carrier, bandwidth, array geometry, and acquisition procedure. Multiple trajectories, sessions, flights, links, or simulation realizations are retained when they provide independent partition units, whereas alternative towns, carrier settings, weather conditions, or array variants from the same source are excluded. This choice broadens mechanism-level diversity without allowing one source to dominate through many closely related configurations.

\begin{figure*}[t]
\centering
\includegraphics[width=\textwidth]{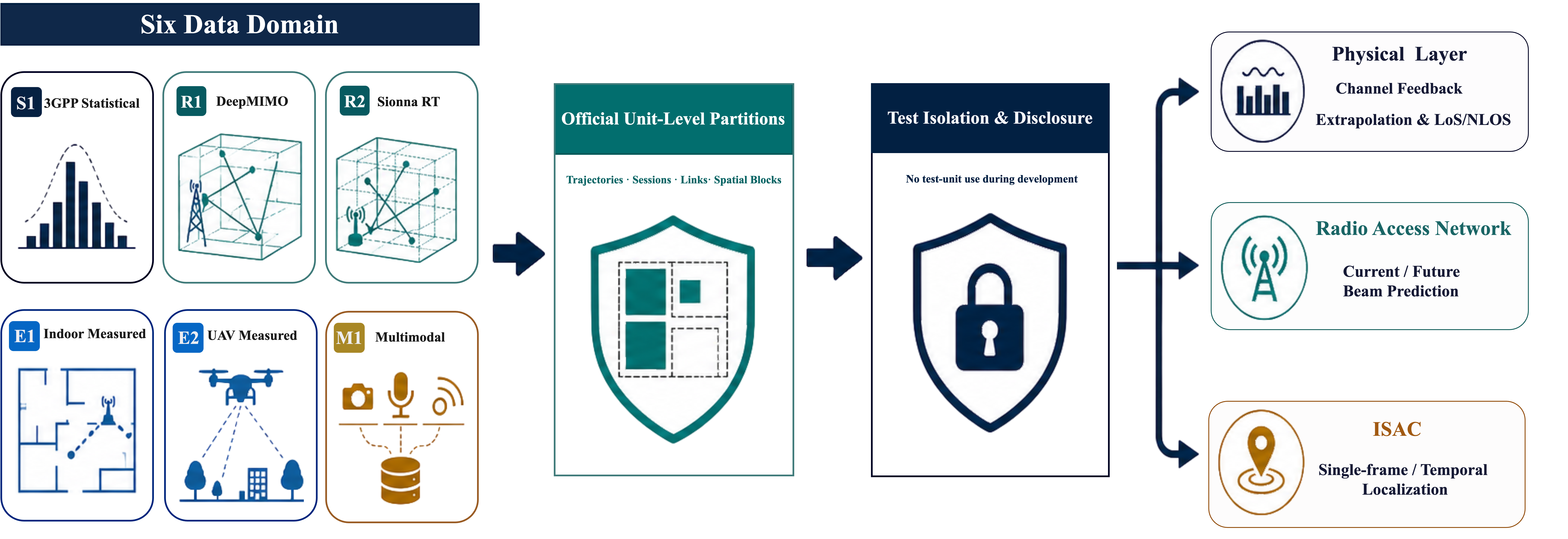}
\caption{CFM-Bench defines independent train, validation, and test units across six data domains, records the data used during model development, and organizes six task groups along PHY, RAN, and ISAC application dimensions.}
\label{fig:overview}
\end{figure*}

Differences in antenna layout, bandwidth, frequency grid, and sampling rate are treated as evaluation dimensions rather than inconsistencies to be erased. Source-aware interfaces preserve the corresponding physical metadata, while model-side resampling, normalization, padding, and tokenization are reported as part of each experimental configuration. As summarized in Fig.~\ref{fig:overview}, these heterogeneous observations enter a common sequence of unit-level partitioning, test isolation, and domain-aware evaluation across PHY, RAN, and ISAC tasks.

\subsection{Multi-Mechanism Domains}
\label{sec:design-domains}

The six curated domains implement the test-conditions principle as a diagnosable stress test rather than a homogeneous enlargement of one propagation source. Table~\ref{tab:domains} summarizes their radio configurations. The identifiers encode provenance rather than quality: S denotes statistical simulation, R ray tracing, E a measured environment, and M synchronized multimodal simulation. Counts refer to the official single-frame view; temporal windows are reported separately.

\begin{table*}[t]
\caption{Data Domains and Radio Configurations in CFM-Bench}
\label{tab:domains}
\centering
\footnotesize
\renewcommand{\arraystretch}{1.18}
\setlength{\tabcolsep}{3.8pt}
\begin{tabularx}{\textwidth}{@{}C{0.50cm} L{2.75cm} L{2.35cm} L{2.95cm} C{2.20cm} Y@{}}
\toprule
\textbf{ID} & \textbf{Data source} & \textbf{Mechanism} & \textbf{Scenario} & \shortstack{\textbf{Carrier /}\\\textbf{bandwidth}} & \textbf{Released CSI view} \\
\midrule
\rowcolor{black!3} S1 & 3GPP TR~38.901 UMi & Statistical (Sionna) & Three-site urban microcell & 3.5~GHz / 100~MHz & 9 links; 64 BS ports; 128 tones \\
R1 & DeepMIMO O1\_60 & RT (Wireless InSite) & Outdoor urban grid & 60~GHz / 400~MHz & 64 BS ports; 256 tones \\
\rowcolor{black!3} R2 & MOCSID & RT (Sionna) & Campus; 10 BSs; pedestrians & Unspecified / 1.92~MHz & 10 BS links; 64 tones on a 30-kHz grid \\
E1 & DICHASUS ADXX & Measurement & Industrial indoor & 3.440~GHz / 50.056~MHz & 64 distributed antennas; 1024 tones \\
\rowcolor{black!3} E2 & MaMIMO-UAV & Measurement & Outdoor air-to-ground & 2.61~GHz / 18~MHz & 64 BS elements; 100 tones \\
M1 & Multimodal-Wireless & Multimodal RT & Town05 CBD V2I & 28~GHz / narrowband & $16\times64$ RSU--vehicle channel matrix \\
\bottomrule
\end{tabularx}
\end{table*}
\subsubsection{Scale and Diversity}
\label{subsec:statistics}

\begin{figure}[t]
\centering
\includegraphics[width=\columnwidth]{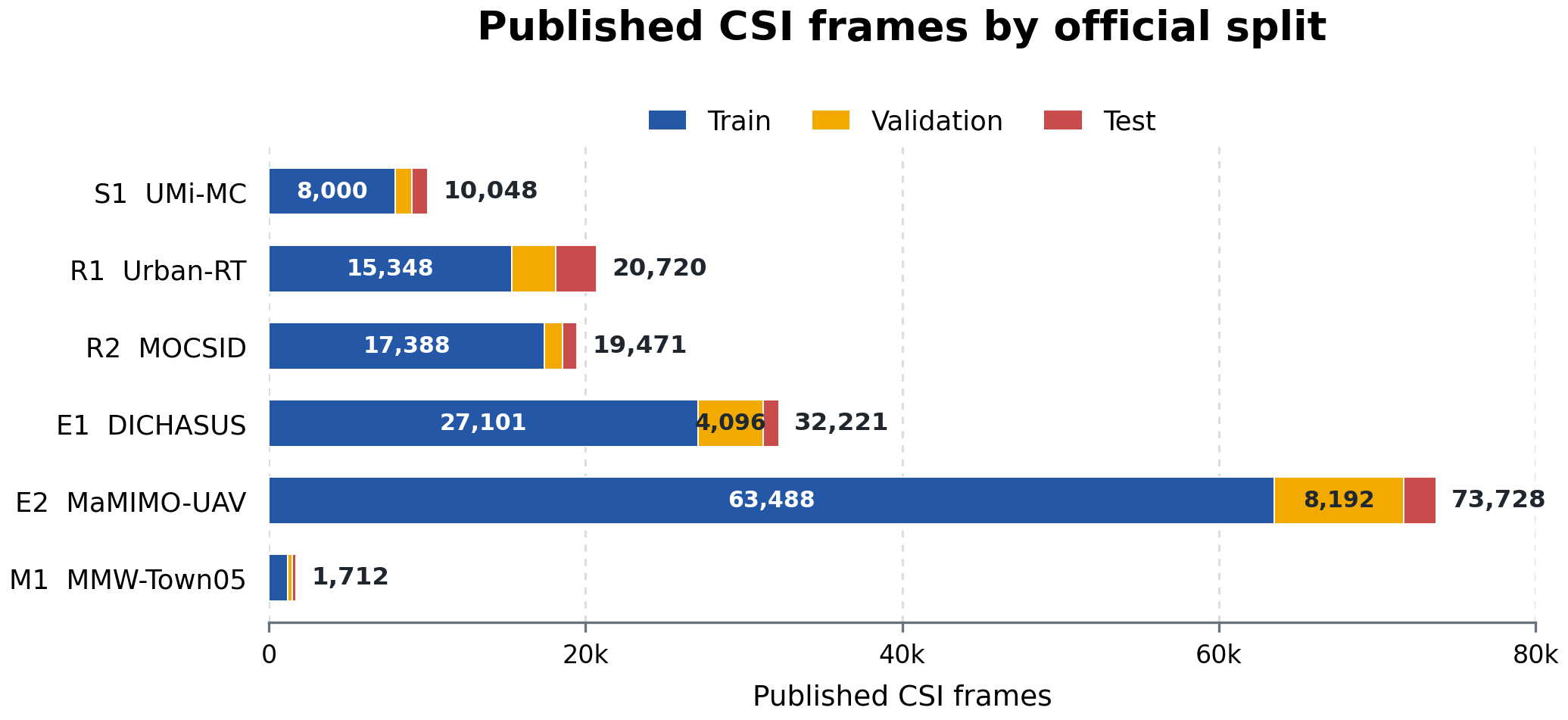}
\caption{Single-frame CSI examples by data domain and split. Bar lengths use a common linear scale; temporal windows are alternative test views and are not added to the totals.}
\label{fig:scale}
\end{figure}

CFM-Bench contains 157,900 single-frame examples: 132,525 for training, 17,523 for validation, and 7,852 for testing. Fig.~\ref{fig:scale} reports these counts by domain and split. The per-domain totals are 10,048 for S1, 20,720 for R1, 19,471 for R2, 32,221 for E1, 73,728 for E2, and 1,712 for M1. The counts refer to benchmark examples retained after quality control and source-specific sampling, rather than every sample in the original datasets. The imbalance is intentional: CFM-Bench preserves independent physical units and reports per-domain scores instead of forcing every source to an arbitrary common size. Temporal windows provide alternative evaluation views of test sequences and are not added to the single-frame totals.

Mechanism diversity does not imply that every target is physically meaningful for every source. Table~\ref{tab:task_support} therefore previews the task coverage obtained after semantic validation and formalized in Section~\ref{sec:protocols}. CSI feedback is available in all six domains, and frequency extrapolation in the five wideband domains, whereas complex temporal extrapolation is restricted to R2 and M1. E1 retains uncompensated measurement phase and E2 does not release a complete future-CSI target view, so neither is included in that task. Current-beam prediction is enabled for S1, R1, R2, E1, and E2, while future-beam prediction is enabled for R2, E1, and E2. M1 provides no official current- or future-beam target. LoS/NLoS supervision is restricted to S1 and R2, and single-frame localization covers all six domains.

\begin{table*}[t]
\caption{Official Downstream-Task Support by Data Domain}
\label{tab:task_support}
\centering
\footnotesize
\renewcommand{\arraystretch}{1.22}
\setlength{\tabcolsep}{3.0pt}
\begin{tabularx}{\textwidth}{@{}L{2.35cm} *{8}{Z}@{}}
\toprule
& \multicolumn{4}{c}{\textbf{PHY}} & \multicolumn{2}{c}{\textbf{RAN}} & \multicolumn{2}{c}{\textbf{ISAC}} \\
\cmidrule(lr){2-5}\cmidrule(lr){6-7}\cmidrule(lr){8-9}
\textbf{Domain} & \shortstack{\textbf{CSI}\\\textbf{feedback}} & \shortstack{\textbf{Frequency}\\\textbf{extrap.}} & \shortstack{\textbf{Temporal}\\\textbf{extrap.}} & \shortstack{\textbf{LoS/}\\\textbf{NLoS}} & \shortstack{\textbf{Current-beam}\\\textbf{prediction}} & \shortstack{\textbf{Future-beam}\\\textbf{prediction}} & \shortstack{\textbf{Frame}\\\textbf{local.}} & \shortstack{\textbf{Temporal}\\\textbf{local.}} \\
\midrule
S1 3GPP UMi & \Supported & \Supported & \Unsupported & \Supported & \Supported & \Unsupported & \Supported & \Unsupported \\
R1 DeepMIMO & \Supported & \Supported & \Unsupported & \Unsupported & \Supported & \Unsupported & \Supported & \Unsupported \\
R2 MOCSID & \Supported & \Supported & \Supported & \Supported & \Supported & \Supported & \Supported & \Supported \\
E1 DICHASUS & \Supported & \Supported & \Unsupported & \Unsupported & \Supported & \Supported & \Supported & \Supported \\
E2 MaMIMO-UAV & \Supported & \Supported & \Unsupported & \Unsupported & \Supported & \Supported & \Supported & \Supported \\
M1 Multimodal & \Supported & \Unsupported & \Supported & \Unsupported & \Unsupported & \Unsupported & \Supported & \Supported \\
\bottomrule
\end{tabularx}
\end{table*}

The diversity attributes in Table~\ref{tab:diversity} complement raw sample count with mechanism and context coverage. The benchmark includes one 3GPP statistical domain, three ray-tracing-based domains, and two measured domains. The deployment settings cover indoor industrial, terrestrial outdoor, aerial, and vehicular links. Four domains preserve temporal sequences; S1 and R2 expose multiple base stations, E1 uses distributed receiving sites, E2 and M1 provide three-dimensional positions, and M1 adds synchronized environmental sensing. The benchmark therefore varies both channel generation and the physical context paired with each observation.

\begin{table*}[t]
\caption{Mechanism-Level and Contextual Diversity of CFM-Bench}
\label{tab:diversity}
\centering
\footnotesize
\renewcommand{\arraystretch}{1.20}
\setlength{\tabcolsep}{2.6pt}
\begin{tabularx}{\textwidth}{@{}L{2.35cm} *{11}{Z}@{}}
\toprule
& \multicolumn{3}{c}{\textbf{Acquisition mechanism}} & \multicolumn{4}{c}{\textbf{Deployment / mobility}} & \multicolumn{4}{c}{\textbf{Structure and context}} \\
\cmidrule(lr){2-4}\cmidrule(lr){5-8}\cmidrule(lr){9-12}
\textbf{Domain} & \textbf{Stat.} & \textbf{RT} & \textbf{Meas.} & \textbf{Indoor} & \shortstack{\textbf{Ground}\\\textbf{outdoor}} & \textbf{Aerial} & \textbf{V2I} & \textbf{Temporal} & \shortstack{\textbf{Multi-site/}\\\textbf{multi-BS}} & \shortstack{\textbf{3D}\\\textbf{position}} & \textbf{Sensors} \\
\midrule
S1 3GPP UMi & \Supported & \Unsupported & \Unsupported & \Unsupported & \Supported & \Unsupported & \Unsupported & \Unsupported & \Supported & \Unsupported & \Unsupported \\
R1 DeepMIMO & \Unsupported & \Supported & \Unsupported & \Unsupported & \Supported & \Unsupported & \Unsupported & \Unsupported & \Unsupported & \Unsupported & \Unsupported \\
R2 MOCSID & \Unsupported & \Supported & \Unsupported & \Unsupported & \Supported & \Unsupported & \Unsupported & \Supported & \Supported & \Unsupported & \Unsupported \\
E1 DICHASUS & \Unsupported & \Unsupported & \Supported & \Supported & \Unsupported & \Unsupported & \Unsupported & \Supported & \Supported & \Unsupported & \Unsupported \\
E2 MaMIMO-UAV & \Unsupported & \Unsupported & \Supported & \Unsupported & \Unsupported & \Supported & \Unsupported & \Supported & \Unsupported & \Supported & \Unsupported \\
M1 Multimodal & \Unsupported & \Supported & \Unsupported & \Unsupported & \Unsupported & \Unsupported & \Supported & \Supported & \Unsupported & \Supported & \Supported \\
\bottomrule
\end{tabularx}
\end{table*}

\subsubsection{Statistical and Ray-Traced Domains}

\textbf{S1: 3GPP urban microcell.} S1 follows the 3GPP TR~38.901 urban-micro street-canyon model~\cite{3gpp38901}, implemented in Sionna with three sites and three sectors per site. Each sector uses an $8\times8$ uniform rectangular array (URA), while the user equipment (UE) has one antenna. The dataset contains 8,000 training, 1,024 validation, and 1,024 test snapshots drawn from mutually disjoint simulation realizations and UE routes. It is treated as static because independently generated channel segments do not provide physically continuous complex CSI. S1 therefore supports frequency extrapolation, LoS/NLoS classification, per-valid-sector current-beam prediction, and single-frame localization, but no temporal task. Each observation retains nine sector links and their validity mask; this is a multi-link tensor, not a single 576-port array.

\textbf{R1: DeepMIMO O1\_60.} R1 uses the 60-GHz DeepMIMO O1\_60 outdoor scenario, whose site-specific propagation paths were generated with Remcom Wireless InSite~\cite{alkhateeb2019deepmimo}. It contains a static two-dimensional user grid, an $8\times8$ base-station URA, and 256 subcarriers over 400 MHz. One of every 50 spatial samples is retained after partitioning to keep the benchmark size tractable. The coverage area is divided into ten non-overlapping regions, with 1-m guard bands preventing adjacent grid points from crossing split boundaries. Samples with no valid ray-traced path are removed. LoS/NLoS classification is not an official R1 task because its held-out regions contain only NLoS examples.

\textbf{R2: MOCSID.} MOCSID models an outdoor campus with ten base stations, overlapping coverage, mixed line-of-sight (LoS) and non-LoS propagation, and spatially consistent pedestrian mobility~\cite{makhlouf2025mocsid,makhlouf2025mocsiddata}. CFM-Bench selects ten complete trajectories while retaining their temporal samples, visible-base-station information, and positions. Frequency-domain CSI is derived from the released path coefficients and delays on an explicitly benchmark-defined 64-tone, 30-kHz relative baseband grid; it is not presented as an upstream OFDM configuration. The per-BS LoS/NLoS labels are geometry-derived and use a link-valid mask. Eight trajectories form the training set, and one trajectory is assigned to each evaluation split.

\subsubsection{Measured Channel Domains}

\textbf{E1: DICHASUS ADXX.} CFM-Bench uses the public DICHASUS ADXX dataset recorded in the ARENA2036 facility~\cite{euchner2021dichasus,euchner2024dichasusadxx}. The source specifies a 3.440-GHz carrier, 50.056-MHz bandwidth, 1024 subcarriers, 64 antennas arranged as four distributed $2\times8$ arrays, and a 48-ms measurement interval. Upstream reference-channel phase/time compensation is not applied in the release. Consequently, beamforming is performed only within each physical array and raw complex temporal CSI extrapolation is disabled, while magnitude-robust future-beam prediction and temporal localization remain supported. Two sessions are assigned to training, and distinct sessions define validation and test; retained windows never cross acquisition gaps.

\textbf{E2: MaMIMO-UAV.} E2 captures a measured air-to-ground channel between a 64-element upward-facing URA and a UAV following three-dimensional trajectories~\cite{cui2023mamimouav,colpaert2023mamimouavdata}. It retains the 2.61-GHz carrier, 18-MHz occupied bandwidth, 100 subcarriers, 1-ms sampling interval, and flight metadata. Two all-zero acquisition segments are removed. Geographic coordinates are interpolated in double precision before conversion to local east--north metres, eliminating quantization steps that would otherwise corrupt millisecond-scale motion. Four flights are assigned to training and one each to validation and test. Complex temporal CSI extrapolation is disabled, but future-beam prediction and temporal localization remain supported on continuity-safe segments.
\subsubsection{Synchronized Multimodal Domain}

\textbf{M1: Multimodal-Wireless.} Multimodal-Wireless uses CARLA and Sionna to pair CSI with environmental sensing~\cite{mao2026multimodalwireless}. CFM-Bench uses the Town05 CBD Crossroad under sunny weather and a 28-GHz vehicle-to-infrastructure link. The narrowband $16\times64$ CSI is aligned at 100 Hz with four vehicle-mounted RGB cameras, LiDAR, and inertial records. Three distinct vehicle links define the training, validation, and test partitions, containing 1,200, 256, and 256 retained frames, respectively. Pose, GPS, and sensor world-coordinate fields are treated as target metadata and excluded from localization inputs to avoid direct target leakage; admissible inputs include CSI and synchronized modalities that do not reveal absolute pose. This protocol tests unseen-link transfer within one scene, not transfer across towns or weather.

\subsection{Leakage-Resistant Partitions and Label Analysis}
\label{sec:design-partition}

The domain descriptions above establish complementary physical conditions. To account for the correlation structure within each condition, CFM-Bench partitions at the largest available independent unit: a complete simulation realization or trajectory, a measurement session, a UAV flight, a vehicle link, or a buffered spatial region. When a source provides enough comparable units, an 8:1:1 allocation is used. When only a few units are available, validation and test each receive a distinct held-out unit, and windows and derived examples remain within that boundary.

Table~\ref{tab:splits} jointly reports dataset scale and the unit of independence. Each dataset row gives single-frame examples followed by the number of source isolation units; the total row sums frames only. Downstream fine-tuning draws from the listed training units, with the complete pretraining and test-isolation methodology specified in Section~\ref{sec:protocols}. The temporal test views contain 64 windows of length 32 for E1, 64 of length 64 for E2, 8 of length 32 for R2, and 16 of length 16 for M1. These windows remain within their respective held-out source units and do not change the single-frame counts.

\begin{table*}[t]
\caption{Dataset Scale and Leakage-Resistant Partitions}
\label{tab:splits}
\centering
\footnotesize
\renewcommand{\arraystretch}{1.18}
\setlength{\tabcolsep}{3.6pt}
\begin{tabularx}{\textwidth}{@{}C{0.50cm} L{2.70cm} C{2.25cm} C{2.25cm} C{2.25cm} Y@{}}
\toprule
\textbf{ID} & \textbf{Partition unit} & \textbf{Train} & \textbf{Validation} & \textbf{Test} & \textbf{Isolation criterion} \\
\midrule
S1 & Simulation realization & 8,000 / 16 & 1,024 / 2 & 1,024 / 2 & Disjoint random seeds and UE routes \\
R1 & Spatial region & 15,348 / 8 & 2,780 / 1 & 2,592 / 1 & 1-m guard bands between splits \\
R2 & Pedestrian trajectory & 17,388 / 8 & 1,175 / 1 & 908 / 1 & Each trajectory belongs to one split \\
E1 & Measurement session & 27,101 / 2 & 4,096 / 1 & 1,024 / 1 & No source session crosses a split \\
E2 & UAV flight & 63,488 / 4 & 8,192 / 1 & 2,048 / 1 & No source flight crosses a split \\
M1 & Vehicle link & 1,200 / 1 & 256 / 1 & 256 / 1 & No source vehicle link crosses a split \\
\midrule
All & -- & 132,525 & 17,523 & 7,852 & 157,900 frames in total \\
\bottomrule
\end{tabularx}
\end{table*}

\begin{figure*}[t]
    \centering
    \includegraphics[width=\textwidth]{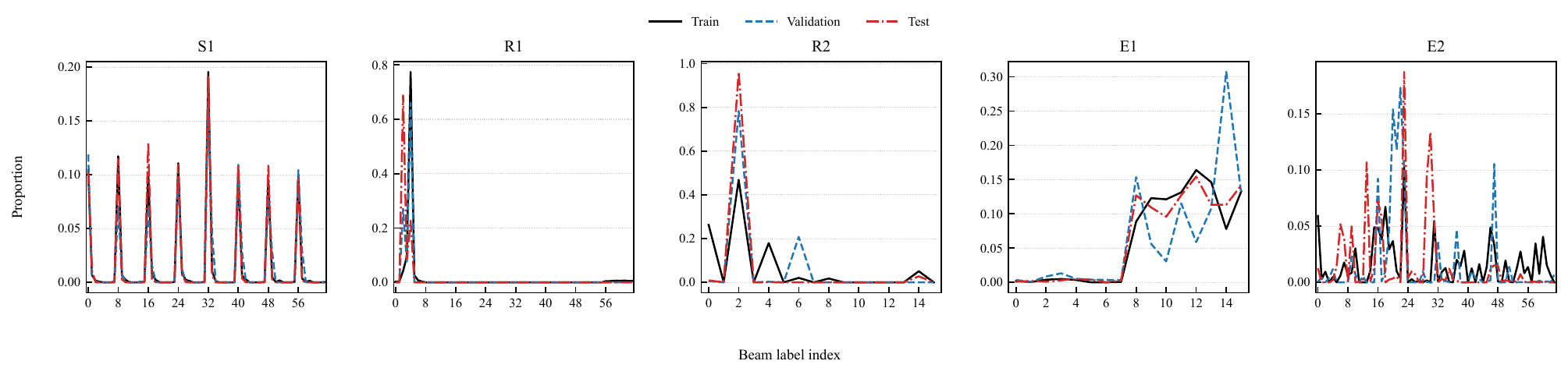}
    \caption{Split-wise distributions of current-beam labels in the five eligible domains. Each curve reports the within-split proportion assigned to each domain-specific beam or beam-pair index; the training, validation, and test distributions are normalized independently. S1, R1, and E2 use 64-candidate codebooks, whereas R2 and E1 use 16-candidate codebooks. Indices are defined within each domain and are not comparable across different codebooks.}
    \label{fig:beam_label_distribution}
\end{figure*}

Fig.~\ref{fig:beam_label_distribution} complements Table~\ref{tab:splits} by showing the current-beam label coverage under the official partitions. The distributions are strongly non-uniform and vary across splits. S1 retains a similar periodic concentration pattern, whereas R1 and R2 are dominated by a small subset of their domain-specific codewords. E1 covers most of its 16-candidate space but exhibits split-dependent changes in label proportions. E2 is distinctive because eight beam indices that are active in the test split do not occur in the training split. These split-dependent label shifts are reported explicitly as part of the official evaluation conditions.

\begin{figure*}[t]
    \centering
    \includegraphics[width=\textwidth]{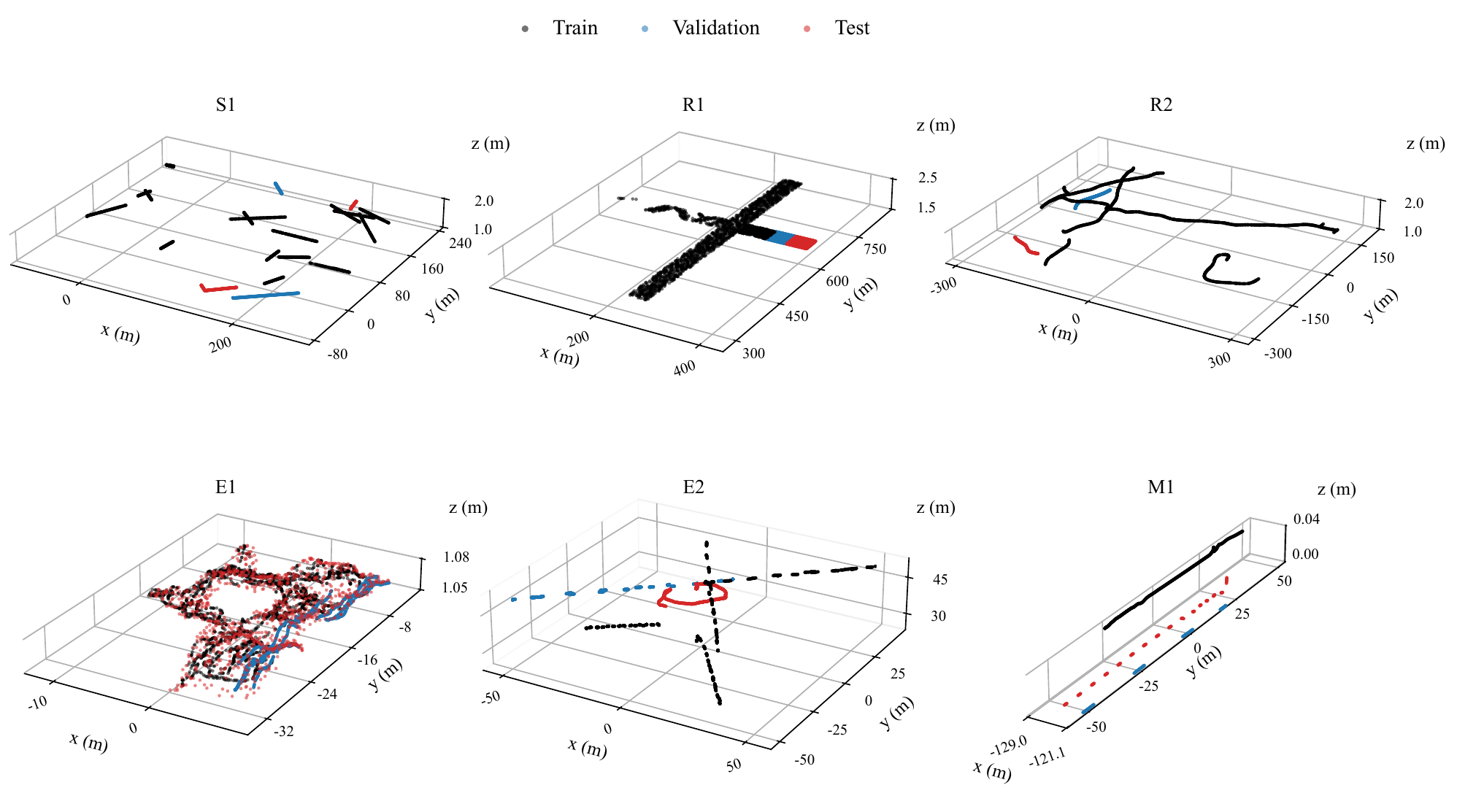}
    \caption{Spatial support of the position labels in the official training, validation, and test splits across the six domains. The plots visualize the domain-specific partition geometry rather than comparable sample densities.}
    \label{fig:position_label_distribution}
\end{figure*}

Fig.~\ref{fig:position_label_distribution} further illustrates the domain-specific partition structures. S1 separates simulation realizations and UE routes, R1 uses buffered spatial regions, R2 isolates complete pedestrian trajectories, E1 separates measurement sessions, E2 isolates UAV flights, and M1 separates vehicle links. Consequently, spatial overlap may remain when independence is defined by acquisition unit rather than location, as in E1.

Taken together, these complementary partition structures enable evaluation under multiple forms of physically meaningful, unit-level generalization. The benchmark therefore evaluates generalization to physically independent evaluation units rather than relying on interpolation among highly correlated channel samples produced by random frame-level splitting.

\subsection{Data Format and Access}
\label{sec:design-access}

The partition definitions establish the evaluation subsets, while the data interface specifies the content of each example. CFM-Bench preserves source-specific CSI structures rather than requiring every domain to share a fixed model input shape. Each example includes complex CSI and the task annotations and physical metadata that are meaningful for its source, such as frequency coordinates, array information, timestamps, positions, base-station identities, and synchronized sensing modalities. The dataset documentation specifies the representation and units for each domain.

Dynamic examples preserve timestamp order, and a temporal window never crosses a trajectory, session, flight, vehicle-link, acquisition gap, or retained-segment boundary. The sensing streams in M1 retain their original frame alignment. Task-specific input definitions exclude target-bearing position information from localization. Resampling, antenna selection, normalization, padding, and tokenization are documented as model-specific choices, and fitted preprocessing statistics are estimated from training data only.

\subsection{Quality Control and Task Eligibility}
\label{subsec:quality}

Quality control is performed at both numerical and semantic levels. Every retained CSI tensor is finite and contains nonzero energy; all-zero R1 spatial samples and two all-zero E2 acquisition segments are removed. R2 may contain an all-zero channel for an individual unavailable BS, but such a link is retained only when it matches the corresponding validity mask. An unavailable link is excluded from link-specific losses and metrics without discarding a frame whose other BS links remain meaningful. Beam labels are recomputed from the CSI and codebooks, and position labels are checked for finite values, units, coordinate references, and continuity within every temporal segment.

Task eligibility follows the data-generating semantics rather than the presence of a convenient tensor axis. S1 is treated as static snapshots because independently generated channel segments do not constitute continuous complex CSI. R1 is a spatial grid rather than a time series. E1 does not expose uncompensated raw complex CSI as a temporal-extrapolation task, and E2 does not provide complete future-CSI supervision. M1 is narrowband and therefore does not support frequency extrapolation. These exclusions do not invalidate other tasks derived from the same observations.

With the data domains, isolation units, and validity boundaries established, the next section specifies the corresponding evaluation methodology and task definitions.

\section{Benchmark Tasks and Evaluation Protocol}
\label{sec:protocols}

\subsection{Fair Comparison and Data-Exposure Reporting}

CFM-Bench does not prescribe an external pretraining corpus, objective, model architecture, or fine-tuning strategy. To distinguish representation pretraining from downstream adaptation, benchmark examples are excluded from foundation-model pretraining. Parameter updates for a benchmark task use the official training split, whereas the validation split supports model and hyperparameter selection without contributing gradient updates. Test trajectories, measurement sessions, vehicle links, spatial blocks, simulator groups, and their derived samples are reserved for final inference and scoring; they are not used to estimate preprocessing statistics or to guide model development.

Each result is accompanied by a data-exposure statement that summarizes the CFM-Bench and external data used during model development. For external pretraining, the statement notes potential overlap in maps, sites, measurement campaigns, simulator scenes, trajectories, flights, sessions, or vehicle links. Results obtained after access to an official test unit, or to unpublished samples from that unit, are reported separately as \emph{test-exposed} or \emph{transductive} rather than grouped with the standard setting. This distinction is important because several domains isolate trajectories, sessions, links, or spatial regions within a shared broader environment. M1, for example, evaluates an unseen vehicle link within the same Town05 run and thus provides unit-level, rather than unseen-scene, generalization.

Comparisons between a CFM and a task-specific model control the available observations and modalities, training examples, validation and test units, and target definition. Parameter count and inference cost are reported with the comparison, together with training compute when available. Scores are presented separately for each domain; an unweighted macro-average across eligible domains serves only as a secondary summary.

\subsection{Common Metric Conventions}

Let $\mathcal{V}_i$ denote the valid target entries for sample $i$ after applying validity and task masks, and let $\mathbf{h}_i$ and $\widehat{\mathbf{h}}_i$ be vectorized complex targets and predictions on $\mathcal{V}_i$. Zero-energy targets are excluded and counted in the evaluation log. The normalized mean-squared error in decibels is
\begin{equation}
\NMSE_{\mathrm{dB}}=10\log_{10}
\frac{\sum_i\|\widehat{\mathbf{h}}_i-\mathbf{h}_i\|_2^2}
{\sum_i\|\mathbf{h}_i\|_2^2}.
\label{eq:nmse}
\end{equation}
The squared generalized cosine similarity is
\begin{equation}
\SGCS=\frac{1}{N}\sum_{i=1}^{N}
\frac{|\mathbf{h}_i^{H}\widehat{\mathbf{h}}_i|^2}
{\|\mathbf{h}_i\|_2^2\|\widehat{\mathbf{h}}_i\|_2^2}.
\label{eq:sgcs}
\end{equation}
NMSE measures complex reconstruction or extrapolation fidelity, whereas SGCS measures directional agreement relevant to spatial processing. Both are common in CSI learning studies~\cite{guo2022overview}.

Table~\ref{tab:tasks} summarizes the taxonomy, eligible domains, targets, and primary evaluation metrics of the six official task groups. Channel extrapolation and localization each expose two evaluation variants, producing the eight support columns in Table~\ref{tab:task_support}; detailed observation budgets and secondary metrics are defined below. An unsupported domain--task combination is omitted rather than assigned a synthetic label.

\begin{table*}[t]
\caption{Task Taxonomy and Primary Evaluation Metrics}
\label{tab:tasks}
\centering
\footnotesize
\renewcommand{\arraystretch}{1.18}
\setlength{\tabcolsep}{4.0pt}
\begin{tabularx}{\textwidth}{@{}C{1.15cm} L{3.10cm} L{4.35cm} L{3.65cm} Y@{}}
\toprule
\textbf{Dim.} & \textbf{Task} & \textbf{Eligible domains} & \textbf{Prediction target} & \textbf{Primary metric} \\
\midrule
\PHYDim & CSI feedback & All six domains & Full CSI from a compressed latent representation & NMSE (dB) \\
\PHYDim & Channel extrapolation & Frequency: S1, R1, R2, E1, E2; temporal: R2, M1 & Missing-frequency or future CSI & Target-only NMSE (dB) \\
\PHYDim & LoS/NLoS classification & S1, R2 & Binary state of each valid link & Macro-F1 \\
\addlinespace[2pt]
\RANDim & Current-beam prediction & S1, R1, R2, E1, E2 & Current codebook-defined beam or beam pair & Top-1/Top-3 accuracy \\
\RANDim & Future-beam prediction & R2, E1, E2 & Future codebook-defined beam or beam pair & Horizon-wise Top-1/Top-3 accuracy \\
\addlinespace[2pt]
\ISACDim & Localization & Single frame: all; temporal: R2, E1, E2, M1 & Native-frame Cartesian position in metres & Mean 3D error (m) \\
\bottomrule
\end{tabularx}
\end{table*}

\subsection{PHY}

\textbf{CSI feedback:} Accurate downlink CSI at the base station (BS) is essential for precoding, beamforming, scheduling, and link adaptation. In frequency-division-duplex massive-MIMO systems, uplink and downlink channels are not directly reciprocal, so the user equipment (UE) must convey a compact representation of its estimated downlink CSI to the BS. Learning-based feedback addresses the resulting dimensionality challenge by jointly training a UE-side encoder and a BS-side decoder, while practical comparisons must also account for encoder complexity, generalization, and encoder--decoder interoperability~\cite{guo2022overview}. CSI feedback enhancement is also a representative two-sided air-interface AI/ML use case studied in 3GPP TR~38.843, where CSI is compressed at the UE and reconstructed at the BS within the legacy feedback framework~\cite{3gpp38843,guo2024ai}.

For sample $i$, let $\mathbf{h}_i\in\mathbb{C}^{D_i}$ contain the released complex CSI coefficients selected by the domain validity mask, and let $\mathbf{x}_i=[\operatorname{Re}(\mathbf{h}_i)^{T},\operatorname{Im}(\mathbf{h}_i)^{T}]^{T}\in\mathbb{R}^{2D_i}$ be their real-valued representation. The encoder $f_{\boldsymbol{\theta}}(\cdot)$ maps $\mathbf{x}_i$ to an $M_i$-dimensional latent representation, and the decoder $g_{\boldsymbol{\phi}}(\cdot)$ reconstructs it:
\begin{equation}
    \mathbf{z}_i=f_{\boldsymbol{\theta}}(\mathbf{x}_i),\qquad
    \widehat{\mathbf{x}}_i=g_{\boldsymbol{\phi}}(\mathbf{z}_i).
\end{equation}
The dimensional compression ratio is $\gamma_i=M_i/(2D_i)$. Within each domain, models are compared at the same declared $\gamma_i$ and over the same valid CSI coefficients. The reconstructed real and imaginary components form $\widehat{\mathbf{h}}_i$. The primary metric is NMSE in decibels as defined in~\eqref{eq:nmse}; lower values indicate more accurate CSI reconstruction. SGCS in~\eqref{eq:sgcs} is secondary and measures preservation of the complex channel direction, with values closer to one indicating stronger similarity. These complementary error and similarity measures are commonly used for CSI feedback and are considered in 3GPP TR~38.843~\cite{guo2022overview,3gpp38843}.

\textbf{Channel extrapolation:} Channel extrapolation involves predicting wireless channel states in new or unseen domains, such as extending limited pilot-based CSI measurements across frequency, time, or space, to support more efficient beamforming and resource management in dynamic 6G environments \cite{gao2026csiextra,SSnet2025gao,gao2025enabling}. Let $\mathbf{H}$ denote the complex channel representation supplied by an eligible domain. In frequency extrapolation, the official index arrays divide the valid subcarriers into central observed bins $\Omega_{\mathrm{obs}}$ and outer target bins $\Omega_{\mathrm{tar}}$. The model receives
\begin{equation}
    \mathbf{H}_{\mathrm{obs}}=\mathbf{M}_{\mathrm{obs}}\odot\mathbf{H}
\end{equation}
together with the indices and allowed physical metadata, and predicts only $\Omega_{\mathrm{tar}}$. M1 is excluded because its CSI is narrowband. In the fixed-horizon temporal-extrapolation protocol, a model receives a causal context window and predicts future complex CSI at frame horizons $h\in\{1,5,10\}$ without using future metadata unavailable online. Target-only NMSE is primary; SGCS and the complete horizon-wise scores are also reported so degradation with prediction distance remains visible.

In addition to the fixed-horizon protocol, CFM-Bench includes a declared sequence-to-sequence variant for models whose native interface jointly predicts a future block. A model receives $L$ consecutive historical frames and predicts the following $P$ frames. The values of $L$ and $P$ are reported and held constant for all methods within a result table. The primary metric is target-only NMSE over all valid coefficients in the $P$ predicted frames, converted to decibels according to~\eqref{eq:nmse}; horizon-wise NMSE and SGCS may be reported as diagnostics. Results obtained using different $(L,P)$ settings are not directly comparable. Under either variant, each window remains within a trajectory, session, flight, vehicle-link, or retained-segment boundary.

\textbf{LoS/NLoS classification:} For S1 and R2, $y_{i,b}\in\{0,1\}$ denotes the NLoS/LoS state of infrastructure link $b$ in frame $i$. A separate link-valid mask excludes unavailable links from both training loss and evaluation. Macro-F1 over valid links is primary because the two propagation states can be imbalanced; accuracy, per-class recall, and the number of evaluated links are secondary. The remaining domains are marked unsupported rather than receiving labels inferred from channel magnitude. This PHY task characterizes propagation state rather than a network control action.

\subsection{RAN}

\textbf{Current-beam prediction:} Beam management refers to the ongoing process of selecting, refining, and tracking the optimal narrow beams in massive MIMO systems to maintain strong links amid user mobility, blockages, and changing environments \cite{jin2026generalizable}. Each eligible domain publishes a current optimal beam or beam-pair target under its fixed codebook. A learned method predicts that target from its declared CSI and/or reference signal received power (RSRP) observations. For a single-sided codebook $\mathcal{F}=\{\mathbf{f}_1,\ldots,\mathbf{f}_J\}$, the target is
\begin{equation}
    j_i^\star=\arg\max_j G_i(j),\qquad
    G_i(j)=\|\mathbf{H}_i\mathbf{f}_j\|_F^2.
\end{equation}
The primary metrics are per-domain Top-$k$ accuracy for $k\in\{1,3\}$. Top-1 measures exact optimal-beam prediction, whereas Top-3 credits a prediction when the optimal beam is among the three highest-ranked candidates. The accuracies are reported separately for each domain rather than pooled across incompatible codebooks. As secondary metrics, normalized beamforming gain is
\begin{equation}
    g_i^{\mathrm{norm}}=\frac{G_i(\widehat{j}_i)}{G_i(j_i^\star)}.
\end{equation}
The corresponding dB gain regret is also reported as a secondary measure; together, these gain-based metrics distinguish physically consequential errors from mistakes between nearly tied beams. Current-beam prediction is evaluated on S1, R1, R2, E1, and E2; M1 contains no official beam target. The reported configuration includes the codebook construction, available input modalities, and temporal context; beam-management terminology follows the broader taxonomy in~\cite{xue2024beammanagement}.

\textbf{Future-beam prediction:} This task keeps the discrete decision target separate from temporal CSI extrapolation. From a four-frame causal observation window, the model predicts the optimal beam or beam-pair index at horizons $h\in\{1,5,10\}$. Horizon-wise Top-1 and Top-3 accuracy are primary, and their arithmetic means across the official horizons provide the corresponding summary scores. Horizon-wise normalized gain and dB gain regret are secondary and expose the physical severity of errors as prediction distance increases. S1 and R1 do not provide temporal ordering, while M1 contains no future-beam target; the task is therefore evaluated on R2, E1, and E2.

\subsection{ISAC}

\textbf{Wireless positioning:} 
Wireless positioning uses radio signal features, such as time of arrival, angle of arrival, or received signal strength, to estimate the physical location of users or devices $\widehat{\mathbf{p}}_i=[x_i,y_i,z_i]^T\in\mathbb{R}^{3}$ in real-world environments \cite{gao2026sidelink,xu2025enhanced}. Mean 3D Euclidean error, i.e., $e_i=\|\widehat{\mathbf{p}}_i-\mathbf{p}_i\|_2$ is primary performance metric; median, 95th-percentile, and horizontal error are secondary. For effectively planar domains, the horizontal metric exposes performance without assuming that independent maps share one coordinate system. The single-frame variant predicts the current position from one aligned observation. The temporal variant maps a causal CSI sequence to the complete aligned position sequence and is available for R2, E1, E2, and M1. Per-domain or per-scene coordinate normalization is estimated from training positions, and predictions are inverse-transformed before scoring in the original metre-valued frame. Coordinates from unrelated maps are evaluated in their respective native frames rather than concatenated into a global regression space. In M1 localization, pose, GPS, true-ego position, and sensor world-coordinate fields are excluded because they directly reveal the target; the evaluated inputs comprise CSI and non-target-bearing synchronized modalities. This task evaluates sensing-oriented spatial inference and does not constitute joint ISAC waveform-design evaluation.

\section{Benchmark Experiments}
\label{sec:experiments}

This section reports representative baseline experiments for four of the six benchmark task groups: CSI feedback, channel extrapolation, current-beam prediction, and wireless positioning. All experiments use the official training, validation, and test partitions and report results separately for each data domain. The evaluated methods follow the data-exposure and test-isolation policy in Section~\ref{sec:protocols}. FLOPs are expressed in millions (M) or billions (G), according to scale. Boldface is determined from the unrounded results.

\subsection{CSI feedback}

We compare three representative CSI-feedback architectures: the convolutional CsiNet~\cite{wen2018csinet}, the lightweight multilayer-perceptron-based M-Net~\cite{yu2023mnet}, and the attention-based TransNet~\cite{cui2022transnet}. The table reports compression factors $\mathrm{CR}=1/\gamma\in\{4,16,64\}$, corresponding to latent fractions $\gamma\in\{1/4,1/16,1/64\}$.

For each domain, the CSI is mapped to its declared angular--delay or angular--angular representation and normalized using statistics estimated from the training split. The UE-side encoder compresses this representation, and the BS-side decoder reconstructs it. The decoded output is then denormalized and inverse-transformed to the native CSI grid before evaluation. Checkpoints are selected using validation NMSE, while the NMSE and SGCS in Table~\ref{tab:csi_feedback} are computed on the held-out test split in the reconstructed full-CSI domain.

\begin{table*}[t]
\centering
\caption{CSI feedback performance and computational complexity across six benchmark datasets. NMSE is reported in dB, SGCS follows~\eqref{eq:sgcs}, and FLOPs use M or G according to scale. The best result for each metric under each compression ratio (CR) is highlighted in bold.}
\label{tab:csi_feedback}
\scriptsize
\setlength{\tabcolsep}{3.5pt}
\renewcommand{\arraystretch}{1.06}

\begin{tabular}{cc|ccc|ccc|ccc}
\toprule
\multirow{2}{*}{CR} &
\multirow{2}{*}{Model} &
\multicolumn{3}{c|}{S1} &
\multicolumn{3}{c|}{R1} &
\multicolumn{3}{c}{R2} \\
\cmidrule(lr){3-5}
\cmidrule(lr){6-8}
\cmidrule(lr){9-11}
& &
NMSE$\downarrow$ & SGCS$\uparrow$ & FLOPs &
NMSE$\downarrow$ & SGCS$\uparrow$ & FLOPs &
NMSE$\downarrow$ & SGCS$\uparrow$ & FLOPs \\
\midrule
\multirow{3}{*}{4}
& CsiNet~\cite{wen2018csinet}
& -11.90 & 0.204 & 31M
& -0.75  & 0.087 & 95M
& -13.26 & 0.762 & 5M \\
& M-Net~\cite{yu2023mnet}
& -16.24 & \textbf{0.528} & 30M
& -0.90  & 0.133 & 97M
& -15.32 & 0.903 & 2M \\
& TransNet~\cite{cui2022transnet}
& \textbf{-16.91} & 0.524 & 170M
& \textbf{-1.08}  & \textbf{0.183} & 387M
& \textbf{-15.48} & \textbf{0.945} & 38M \\
\midrule
\multirow{3}{*}{16}
& CsiNet
& -11.69 & 0.327 & 18M
& -0.39  & 0.029 & 45M
& -11.41 & 0.637 & 4M \\
& M-Net
& -13.83 & 0.362 & 17M
& -0.41  & 0.102 & 47M
& -14.33 & 0.824 & 2M \\
& TransNet
& \textbf{-14.68} & \textbf{0.437} & 158M
& \textbf{-0.93}  & \textbf{0.108} & 336M
& \textbf{-15.20} & \textbf{0.924} & 38M \\
\midrule
\multirow{3}{*}{64}
& CsiNet
& -8.62  & 0.122 & 15M
& 0.20   & 0.006 & 33M
& -6.20  & 0.346 & 4M \\
& M-Net
& -9.52  & 0.180 & 14M
& -0.13  & 0.051 & 34M
& -11.14 & 0.623 & 1M \\
& TransNet
& \textbf{-10.56} & \textbf{0.237} & 155M
& \textbf{-0.78}  & \textbf{0.073} & 324M
& \textbf{-13.03} & \textbf{0.864} & 37M \\
\bottomrule
\end{tabular}

\vspace{1.0mm}

\begin{tabular}{cc|ccc|ccc|ccc}
\toprule
\multirow{2}{*}{CR} &
\multirow{2}{*}{Model} &
\multicolumn{3}{c|}{E1} &
\multicolumn{3}{c|}{E2} &
\multicolumn{3}{c}{M1} \\
\cmidrule(lr){3-5}
\cmidrule(lr){6-8}
\cmidrule(lr){9-11}
& &
NMSE$\downarrow$ & SGCS$\uparrow$ & FLOPs &
NMSE$\downarrow$ & SGCS$\uparrow$ & FLOPs &
NMSE$\downarrow$ & SGCS$\uparrow$ & FLOPs \\
\midrule
\multirow{3}{*}{4}
& CsiNet
& -0.21 & 0.043 & 11M
& 0.54  & 0.015 & 31M
& 0.20  & 0.004 & 11M \\
& M-Net
& -0.20 & 0.041 & 8M
& -0.18 & 0.095 & 30M
& -0.26 & 0.132 & 8M \\
& TransNet
& \textbf{-0.26} & \textbf{0.056} & 79M
& \textbf{-0.70}  & \textbf{0.168} & 170M
& \textbf{-1.34}  & \textbf{0.346} & 79M \\
\midrule
\multirow{3}{*}{16}
& CsiNet
& -0.15 & 0.027 & 8M
& 0.39  & 0.014 & 18M
& 0.13  & 0.003 & 8M \\
& M-Net
& -0.15 & 0.026 & 5M
& 0.26  & 0.015 & 17M
& -0.12 & 0.092 & 5M \\
& TransNet
& \textbf{-0.21} & \textbf{0.041} & 76M
& \textbf{-0.20} & \textbf{0.047} & 158M
& \textbf{-0.59} & \textbf{0.228} & 76M \\
\midrule
\multirow{3}{*}{64}
& CsiNet
& -0.07 & 0.012 & 7M
& 0.25  & \textbf{0.006} & 15M
& 0.04  & 0.002 & 7M \\
& M-Net
& -0.11 & 0.017 & 4M
& 0.13  & 0.004 & 14M
& 0.08  & 0.025 & 4M \\
& TransNet
& \textbf{-0.17} & \textbf{0.029} & 75M
& \textbf{-0.01} & 0.005 & 155M
& \textbf{-0.02} & \textbf{0.097} & 75M \\
\bottomrule
\end{tabular}
\end{table*}

Across the 18 dataset--compression-factor combinations in Table~\ref{tab:csi_feedback}, TransNet achieves the lowest NMSE in every case and the highest SGCS in 16 cases. The few SGCS exceptions indicate that NMSE and SGCS capture complementary aspects of reconstruction fidelity. Increasing the compression factor generally reduces reconstruction quality, although the trend is not monotonic for every model--domain pair.

The observed scores vary substantially across domains. At $\mathrm{CR}=64$, for example, the best NMSE is $-10.56$~dB on S1 and $-13.03$~dB on R2, compared with $-0.78$~dB on R1 and $-0.17$~dB on E1. E2 and M1 also approach 0~dB at this compression factor. Because the domains differ in native CSI shape, energy distribution, preprocessing transform, and partition unit, these values characterize the complete evaluation configuration and do not isolate any single source of the performance differences.

TransNet incurs the highest FLOPs in every evaluated configuration, while CsiNet and M-Net are considerably lighter. The table therefore provides both accuracy and inference-cost reference points without treating computational scale as an explanation for the observed domain-wise rankings.

\subsection{Channel extrapolation}

We evaluate three pretrained channel-prediction baselines: LLM4CP~\cite{liu2024llm4cp}, WiFo~\cite{liu2025wifo}, and CSI-MAE~\cite{jiang2026csimae}. CSI-MAE* augments the released CSI-MAE backbone with frame-wise spatial encoding followed by cross-frame temporal self-attention and rotary positional encoding for time-domain extrapolation. Each baseline is separately fine-tuned for up to 200 epochs on the official training split of each data domain, with early stopping.

For frequency-domain extrapolation, the target frequency bins are masked and reconstructed from the observed bins. For time-domain extrapolation, the experiments use the declared sequence-to-sequence variant with $L=P=5$: five historical CSI frames are used to predict the following five frames. Training and evaluation windows are generated with a stride of one and are discarded if they cross a source-unit boundary or temporal discontinuity. The reported score is NMSE over all valid coefficients in the five predicted frames.

LLM4CP initializes its backbone with pretrained GPT-2 weights and freezes its main pretrained modules, whereas WiFo and CSI-MAE are initialized from their released checkpoints and fine-tuned end to end. When the released patch configuration is incompatible with a target CSI tensor, the corresponding patch-dependent layers of WiFo and CSI-MAE are reinitialized. Model-specific normalization uses only observable CSI: WiFo and CSI-MAE use batch-wise min--max normalization, while LLM4CP uses sample-wise $z$-score normalization. Predictions are inverse-transformed before scoring on the original complex-valued CSI scale. Because M1 lacks the frequency axis expected by WiFo and CSI-MAE, one antenna-domain dimension is supplied to these models as their nominal frequency dimension to preserve the released architectures with minimal modification. This is an implementation-specific dimensional remapping rather than a physically frequency-selective representation; therefore, the corresponding M1 results should be interpreted with this limitation.

\begin{table}[t]
\centering
\caption{Frequency-domain channel extrapolation performance and computational complexity. Target-only NMSE is reported in dB, FLOPs use M or G according to scale, and the best NMSE on each dataset is highlighted in bold.}
\label{tab:frequency_extrapolation}
\scriptsize
\setlength{\tabcolsep}{3.0pt}
\renewcommand{\arraystretch}{1.10}

\begin{tabular}{c|cc|cc|cc}
\toprule
\multirow{2}{*}{Dataset} & \multicolumn{2}{c|}{LLM4CP~\cite{liu2024llm4cp}} & \multicolumn{2}{c|}{WiFo~\cite{liu2025wifo}} & \multicolumn{2}{c}{CSI-MAE~\cite{jiang2026csimae}} \\
\cmidrule(lr){2-3}\cmidrule(lr){4-5}\cmidrule(lr){6-7}
& NMSE & FLOPs & NMSE & FLOPs & NMSE & FLOPs \\
\midrule
S1 & -1.81 & 49.85G & \textbf{-10.08} & 27.20G & -9.69 & 326.29G \\
R1 & 0.00 & 21.07G & \textbf{-2.60} & 65.14G & -0.99 & 60.47G \\
R2 & -10.46 & 8.90G & \textbf{-18.93} & 2.81G & -6.56 & 37.06G \\
E1 & 0.00 & 88.23G & \textbf{-3.86} & 518.28G & 0.00 & 283.08G \\
E2 & 1.32 & 8.48G & 0.00 & 20.34G & \textbf{-0.04} & 25.72G \\
\bottomrule
\end{tabular}
\end{table}

\begin{table}[t]
\centering
\caption{Time-domain channel extrapolation performance and computational complexity. NMSE over the five predicted frames is reported in dB, FLOPs use M or G according to scale, and the best NMSE on each dataset is highlighted in bold.}
\label{tab:time_extrapolation}
\scriptsize
\setlength{\tabcolsep}{3.0pt}
\renewcommand{\arraystretch}{1.10}

\begin{tabular}{c|cc|cc|cc}
\toprule
\multirow{2}{*}{Dataset} & \multicolumn{2}{c|}{LLM4CP~\cite{liu2024llm4cp}} & \multicolumn{2}{c|}{WiFo~\cite{liu2025wifo}} & \multicolumn{2}{c}{CSI-MAE*~\cite{jiang2026csimae}} \\
\cmidrule(lr){2-3}\cmidrule(lr){4-5}\cmidrule(lr){6-7}
& NMSE & FLOPs & NMSE & FLOPs & NMSE & FLOPs \\
\midrule
M1 & \textbf{-0.50} & 811M & 0.07 & 35.68G & 0.14 & 21.42G \\
R2 & -6.13 & 811M & \textbf{-9.27} & 35.68G & -7.09 & 21.42G \\
\bottomrule
\end{tabular}
\end{table}

Tables~\ref{tab:frequency_extrapolation} and~\ref{tab:time_extrapolation} show that no baseline consistently dominates across all evaluated domains and extrapolation variants. After domain-specific fine-tuning, WiFo exhibits the most consistent performance, achieving the lowest NMSE on four of the five frequency-extrapolation domains and on the R2 time-domain task. CSI-MAE and LLM4CP exhibit stronger domain dependence. CSI-MAE remains close to WiFo on S1, while the temporally extended CSI-MAE* ranks second on the R2 time-domain extrapolation task even though temporal modeling is not part of the original CSI-MAE architecture. LLM4CP does not achieve the lowest NMSE on any frequency-extrapolation domain. It achieves the best result on the M1 time-domain task; nevertheless, all three methods remain near 0~dB on M1.

The FLOPs reported in Tables~\ref{tab:frequency_extrapolation} and~\ref{tab:time_extrapolation} show no consistent association between computational cost and extrapolation accuracy. The resulting performance--complexity trade-off is therefore domain- and task-dependent rather than a uniform property of any one predictor.

E1, E2, and M1 expose distinct evaluation boundaries. E1 produces clear performance differences among the baselines, indicating that the frequency-extrapolation task is learnable under the benchmark protocol while remaining discriminative among the evaluated adaptation pipelines. Although all three baselines attain substantially lower NMSE on the E2 training flights, their validation and test NMSE values remain around 0~dB on the held-out flights. This contrast shows that the task can be fitted within the observed flights but remains challenging under cross-flight generalization. For the MAE-based baselines on M1, the results jointly reflect the difficulty of cross-link temporal prediction and the limitations of adapting pretrained representations to a mismatched narrowband channel configuration.

\subsection{Current-beam prediction}

We compare three task-specific models, namely ResNet50~\cite{he2016deep}, the CNN--self-attention BPNN~\cite{xue2025integrated}, and the position-feature-fusion PFNet~\cite{jin2026generalizable}, with two pretrained CFMs, CSI-MAE~\cite{jiang2026csimae} and LWM~\cite{alikhani2025lwm}. Each model maps a single CSI snapshot to the instantaneous optimal-beam index after the input is min--max normalized to $[-1,1]$. R2 and E1 use 16-beam codebooks, while S1, R1, and E2 use 64-beam codebooks. Owing to space constraints, the main table presents Top-1 accuracy only; Top-3 accuracy remains part of the evaluation protocol defined in Section~\ref{sec:protocols}.

PFNet is trained with the beam index as the primary classification target and position coordinates as auxiliary regression targets. ResNet50 and BPNN do not use this auxiliary position supervision. PFNet requires only CSI at inference, but its training-time access to position labels constitutes additional supervision and is marked by $\dagger$ in the table. Consequently, differences involving PFNet reflect both the learning architecture and the supervision setting and should not be attributed solely to a stronger network structure.

\begin{table*}[t]
\centering
\caption{Current-beam prediction performance and computational complexity. The main table reports Top-1 accuracy owing to space constraints; $\dagger$ denotes additional position-label supervision during training. FLOPs use M or G according to scale, and the best Top-1 result on each dataset is highlighted in bold.}
\label{tab:current_beam_prediction}
\scriptsize
\setlength{\tabcolsep}{2.5pt}
\renewcommand{\arraystretch}{1.10}

\begin{tabular}{c|cc|cc|cc|cc|cc}
\toprule
\multirow{2}{*}{Dataset} & \multicolumn{2}{c|}{ResNet50~\cite{he2016deep}} & \multicolumn{2}{c|}{BPNN~\cite{xue2025integrated}} & \multicolumn{2}{c|}{PFNet$^\dagger$~\cite{jin2026generalizable}} & \multicolumn{2}{c|}{CSI-MAE~\cite{jiang2026csimae}} & \multicolumn{2}{c}{LWM~\cite{alikhani2025lwm}} \\
\cmidrule(lr){2-3}\cmidrule(lr){4-5}\cmidrule(lr){6-7}\cmidrule(lr){8-9}\cmidrule(lr){10-11}
& Top-1 & FLOPs & Top-1 & FLOPs & Top-1 & FLOPs & Top-1 & FLOPs & Top-1 & FLOPs \\
\midrule
S1 & 72.42\% & 1.32G & 74.00\% & 614M & \textbf{78.87\%} & 832M & 24.31\% & 22.55G & 53.99\% & 4.04G \\
R1 & 37.11\% & 2.64G & 23.19\% & 1.23G & \textbf{71.10\%} & 1.66G & 22.36\% & 46.14G & 22.72\% & 4.04G \\
R2 & 97.05\% & 194M & 96.01\% & 77M & \textbf{97.61\%} & 128M & 87.79\% & 2.90G & 93.53\% & 333M \\
E1 & 43.24\% & 10.57G & \textbf{49.49\%} & 4.91G & 48.63\% & 6.65G & 19.34\% & 46.14G & 45.61\% & 4.04G \\
E2 & 7.96\% & 1.14G & 8.11\% & 480M & 1.22\% & 732M & \textbf{18.56\%} & 19.69G & 10.30\% & 2.88G \\
\bottomrule
\end{tabular}
\end{table*}

Table~\ref{tab:current_beam_prediction} shows that no model leads across all five domains. PFNet obtains the highest Top-1 accuracy on S1, R1, and R2, BPNN ranks first on E1, and CSI-MAE ranks first on E2. The two CFMs do not exhibit a uniform advantage over task-specific models: CSI-MAE is strongest on E2 but substantially lower on S1, R1, and E1, while LWM is competitive on E1 and R2 without attaining the best result on either domain.

On S1, the three task-specific models achieve 72.42\%--78.87\%, whereas the two CFMs attain 24.31\% and 53.99\%. R1 produces the largest observed separation: PFNet reaches 71.10\%, compared with 22.36\%--37.11\% for the other methods. Because PFNet alone receives auxiliary position labels among the task-specific baselines, this gap cannot be used as evidence of architectural superiority in isolation. On R2, all five methods exceed 87\%, and the three task-specific models exceed 96\%, leaving comparatively limited separation under the 16-beam codebook.

On E1, BPNN, PFNet, and LWM obtain similar Top-1 accuracies of 49.49\%, 48.63\%, and 45.61\%, respectively. E2 yields lower accuracy for every method, but CSI-MAE reaches 18.56\%, compared with 10.30\% for LWM and at most 8.11\% for the task-specific baselines. E2 combines a held-out-flight split with beam indices that are absent from the training split; the result therefore characterizes performance under this particular acquisition-unit and label-distribution shift. Across the five domains, the rankings vary with the data source, codebook, supervision setting, and adaptation method. BPNN has the lowest FLOPs in every domain, whereas the CFM baselines generally incur higher inference cost without a consistent accuracy advantage.

\subsection{Wireless positioning}

We evaluate three CSI-based positioning baselines. AAResCNN uses attention-augmented residual convolutional blocks~\cite{10018917}, ABPN combines spatial and channel attention~\cite{zhang2024deep}, and DyLoc uses an angle--delay-profile representation for channel-based localization~\cite{9488913}. All methods use one CSI snapshot without temporal context, position metadata, or synchronized non-CSI sensing inputs. Both CSI inputs and coordinate labels are $z$-score standardized using statistics computed exclusively from the training split, and predictions are inverse-transformed to the native coordinate system before scoring.

Table~\ref{tab:wireless_positioning} reports the primary mean 3-D Euclidean error; the median, 95th-percentile, and horizontal errors defined in Section~\ref{sec:protocols} are omitted from the main table for space. The experiment therefore covers the single-frame localization variant only. Evaluating the temporal variant additionally requires a causal sequence model, a common observation length, and continuity-safe windows on R2, E1, E2, and M1.

\begin{table}[t]
\centering
\caption{Single-frame wireless positioning performance and computational complexity. Mean 3-D Euclidean error is reported in metres; FLOPs use M or G according to scale. The lowest mean error on each dataset is highlighted in bold.}
\label{tab:wireless_positioning}
\scriptsize
\setlength{\tabcolsep}{4.0pt}
\renewcommand{\arraystretch}{1.10}

\begin{tabular}{c|cc|cc|cc}
\toprule
\multirow{2}{*}{Dataset} & \multicolumn{2}{c|}{AAResCNN~\cite{10018917}} & \multicolumn{2}{c|}{ABPN~\cite{zhang2024deep}} & \multicolumn{2}{c}{DyLoc~\cite{9488913}} \\
\cmidrule(lr){2-3}\cmidrule(lr){4-5}\cmidrule(lr){6-7}
& Error & FLOPs & Error & FLOPs & Error & FLOPs \\
\midrule
E1 & 2.64 & 2.03G & \textbf{1.34} & 16.15G & 1.61 & 52.57G \\
E2 & 34.20 & 2.03G & 16.37 & 1.58G & \textbf{7.51} & 4.97G \\
M1 & 32.39 & 2.03G & \textbf{22.34} & 252M & 31.36 & 796M \\
R1 & 188.83 & 2.03G & 163.22 & 4.04G & \textbf{131.12} & 13.14G \\
R2 & \textbf{82.74} & 2.03G & 83.17 & 2.52G & 343.95 & 801M \\
S1 & \textbf{64.17} & 2.03G & 92.04 & 18.17G & 188.66 & 6.63G \\
\bottomrule
\end{tabular}
\end{table}

Table~\ref{tab:wireless_positioning} shows that no baseline consistently achieves the lowest positioning error across all six datasets. AAResCNN obtains the lowest observed error on R2 and S1, DyLoc ranks first on E2 and R1, and ABPN performs best on E1 and M1. On R2, AAResCNN and ABPN yield numerically close errors of 82.74 and 83.17~m, respectively. Overall, the changing model rankings demonstrate domain-dependent behavior rather than a uniform architectural advantage.

The per-domain errors range from 1.34--2.64~m on E1 and from 7.51--34.20~m on E2. The lowest observed errors on M1, R1, R2, and S1 are 22.34, 131.12, 82.74, and 64.17~m, respectively. R2 also exhibits the largest within-domain spread, from 82.74 to 343.95~m. Because the domains use different native coordinate systems, spatial extents, and partition units, these values characterize model performance within each domain and do not define a direct cross-domain difficulty ranking.

The complexity results in Table~\ref{tab:wireless_positioning} exhibit no monotonic relationship between FLOPs and positioning error. Lightweight models can be competitive or even optimal on some domains, whereas higher computational cost provides advantages only in specific settings, reinforcing the need for per-domain model selection.

\section{Data Availability and Licensing}
\label{sec:availability}

CFM-Bench provides the six data domains, split definitions, task annotations, evaluation software, quality-control documentation, and licensing notices. The dataset is publicly available online.\footnote{CFM-Bench dataset download: \url{https://www.chaspark.com/\#/s/CFM-Bench}.} For every domain, the documentation identifies the source, independent partition units, retained and held-out sample counts, and benchmark transformations. R1 spatially subsamples the DeepMIMO O1\_60 scenario.

CFM-Bench does not impose one license on heterogeneous upstream sources. S1, E1, and M1 are distributed under CC BY 4.0; E2 is under CC BY-NC 4.0; the R2 derived database retains ODbL 1.0 obligations; and the R1 adaptation uses CC BY-NC-SA 4.0 subject to any separate terms governing the DeepMIMO O1\_60 scenario and underlying ray-tracing assets. Dataset-specific attribution, source links, modification notices, and license texts accompany each domain, while CFM-Bench code and original documentation are released under the MIT License. The benchmark does not replace or override upstream terms, and R1 data are redistributed only where the scenario terms permit it.

\section{Limitations}
\label{sec:limitations}

CFM-Bench deliberately selects breadth across sources instead of exhaustive coverage within each source. One fixed configuration cannot characterize every carrier, antenna, weather condition, or deployment. Broader cross-configuration and out-of-distribution coverage will require additional independently partitionable data in future releases.

The six domains are not statistically exchangeable. Their sample counts, amplitude calibration, label density, hardware effects, propagation assumptions, and split difficulty differ, and a domain identifier can be correlated with carrier and environment. Absolute CSI power is therefore not directly comparable across sources. Scores should be inspected per domain; a macro-average is only a concise summary and not evidence that all sources contribute equivalent difficulty. Similarly, simulated and measured channels serve different purposes, and strong simulation performance alone does not establish robustness to calibration errors or measurement noise.

The public test sets favor reproducibility but cannot prevent undeclared reuse or repeated manual adaptation. A later leaderboard may add hidden trajectories or sessions while retaining the public protocol for local development. Several domains isolate units within a shared environment: E1 emphasizes cross-session robustness, E2 cross-flight transfer, and M1 an unseen vehicle link within one Town05 run rather than unseen-scene generalization. Localization difficulty and effective motion dimensionality consequently differ across domains even though all targets are expressed in metres.

Beam candidate spaces are also heterogeneous. R2 and E1 include infrastructure selection, and E2 contains training-unseen test codewords. Top-1 and Top-3 accuracy are therefore reported per domain as the primary metrics rather than pooled across codebooks; secondary normalized achieved gain and dB gain regret indicate whether a classification error selects a beam that is physically much worse or nearly tied with the oracle. M1 provides no official current- or future-beam target. E2 future-beam horizons are only 1--10~ms and admit a strong persistence baseline. E1 raw complex CSI lacks reference-channel compensation, R2 LoS/NLoS labels are geometry-derived, and the release does not yet provide dual-polarized channel configurations. Finally, CSI-feedback representations are constructed from reference CSI. This protocol measures reconstruction under dimensional compression but does not reproduce every hardware impairment or the complete acquisition and control chain of a deployed FDD system. Heterogeneous antenna and frequency grids also burden architectures that assume a fixed tensor shape; reported performance reflects both representation quality and the declared adaptation strategy.

\section{Conclusion}
\label{sec:conclusion}

This paper introduced CFM-Bench, a unified multi-domain, multi-task benchmark that moves CFM evaluation beyond within-pipeline demonstrations of pretraining gains. Its 157,900 official single-frame examples combine statistical simulation, multiple ray-tracing mechanisms, measured terrestrial and aerial channels, and synchronized vehicular sensing. Unit-level partitions, transparent data-exposure reporting, domain-specific physical semantics, and six task groups, including end-to-end CSI feedback evaluated by NMSE and squared generalized cosine similarity (SGCS), enable reproducible comparison without prescribing a particular CFM pretraining or adaptation strategy. The reference experiments show that model rankings vary across domains and tasks: no CFM or task-specific architecture is uniformly strongest, and comparisons can also depend on downstream supervision, as illustrated by position-assisted beam prediction. CFM-Bench therefore provides a common substrate for studying transfer across models, environments, configurations, and applications under explicitly reported evaluation conditions.

\bibliographystyle{IEEEtran}
\bibliography{cfm_bench}

\end{document}